%% file: main.tex
\documentclass[10pt,twocolumn,letterpaper]{article}

\usepackage{iccv}
\usepackage{times}
\usepackage{epsfig}
\usepackage{graphicx}
\usepackage{amsmath}
\usepackage{amssymb}
\input{newcommands2}

\usepackage[breaklinks=true,bookmarks=false]{hyperref}

\iccvfinalcopy % *** Uncomment this line for the final submission

\def\iccvPaperID{****} % *** Enter the ICCV Paper ID here

\begin{document}

%%%%%%%%% TITLE
\title{3DPeople: Modeling the Geometry of Dressed Humans}

\author{A. Pumarola$^{1}$ \quad
    J. Sanchez$^{1}$ \quad
    G. P. T. Choi$^{2}$ \quad
    A. Sanfeliu$^{1}$ \quad
    F. Moreno-Noguer$^{1}$\\%
$^1$Institut de Rob\`otica i Inform\`atica Industrial, CSIC-UPC\\%
$^2$John A. Paulson School of Engineering and Applied Sciences, Harvard University\\%
}

\maketitle
%\thispagestyle{empty}

%%%%%%%%% ABSTRACT
\begin{abstract}
  \vspace{-2mm}
Recent advances in  3D human shape estimation build upon parametric representations that model very well the  shape of the naked body, but are not appropriate to  represent the clothing geometry. In this paper, we present an approach to model dressed humans and predict their geometry from single images. We contribute in three fundamental aspects of the problem, namely, a new dataset, a novel shape parameterization algorithm and an end-to-end deep generative network for predicting shape.

First, we present {\em 3DPeople}, a large-scale synthetic  dataset  with 2.5 Million  photo-realistic images of 80 subjects  performing 70  activities and wearing  diverse  outfits.  Besides providing textured 3D meshes for clothes and body, we annotate the dataset with segmentation masks, skeletons, depth, normal maps and optical flow. All this together makes  3DPeople  suitable for a plethora of tasks. 

We then represent the 3D shapes using 2D geometry images. To build these images we propose a novel  spherical area-preserving parameterization algorithm based on  the optimal mass transportation method. We show this approach to improve existing spherical maps which tend to shrink the elongated parts of the full body models such as the arms and legs, making the geometry images incomplete. 

Finally, we design a multi-resolution deep generative network that, given an input  image of a dressed human, predicts his/her geometry image (and thus the clothed body shape) in an end-to-end manner. We obtain very promising results in jointly capturing body pose and clothing shape, both for  synthetic validation and on the wild images. 
\end{abstract}

%%%%%%%%% BODY TEXT
\section{Introduction}
With the advent of deep learning, the problem of predicting the geometry of the human body from single images has experienced a tremendous boost. The combination of Convolutional Neural Networks with large  MoCap  datasets~\cite{SigalIJCV2010,IonescuPAMI2014},  resulted in a substantial  number of works that robustly predict the 3D position of the body  joints~\cite{MartinezICCV2017,MehtaSIGGRAPH2017,Moreno_cvpr2017,PavlakosCVPR2017,RogezNIPS2016,SunECCV2018,TomeCVPR2017,VarolCVPR2017,YangCVPR2018}.

In order to  estimate the full body shape a standard practice adopted   in~\cite{BogoECCV2016,DibraCVPR2017,GuanICCV2009,KanazawaCVPR2018,TungNIPS2017,zanfir2018deep}   is to regress the parameters of  low rank parametric models~\cite{Anguelov_SIGGRAPH2005,Loper_SIGGRAPH2015}.  Nevertheless, while these parametric models describe very accurately the geometry of the naked body, they are not appropriate to capture the shape of clothed humans.   

Current trends  focus on proposing alternative representations to the low rank models. Varol \etal~\cite{VarolECCV2018} advocate for a  direct inference of volumetric body shape, although still without accounting for the clothing geometry. Very recently,~\cite{NatsumeCVPR2019} uses 2D silhouettes and the visual hull algorithm to recover shape and texture of clothed human bodies. Despite  very promising results, this approach still requires frontal-view input images of the person with no background, and under relatively simple body poses.

In  this paper, we introduce a general pipeline to estimate the geometry of dressed humans which is able cope with  a wide spectrum of clothing outfits and textures, complex body poses and shapes,  and changing backgrounds and camera viewpoints. For this purpose, we contribute in three key areas  of the problem, namely,  the data collection,  the shape representation and  the image-to-shape inference. 

Concretely, we first present 3DPeople a new large-scale dataset with 2.5 Million photorealistic synthetic images of people under varying clothes and apparel. We split the dataset  40 male/40 female with different body shapes and skin tones, performing 70 distinct actions~(see Fig.~1). The dataset contains  the 3D geometry of both the naked and dressed body, and additional annotations including skeletons, depth and normal maps, optical flow and semantic segmentation masks. This additional data is indeed very similar to SURREAL~\cite{VarolCVPR2017} which was built for similar purposes. The key  difference between SURREAL and 3DPeople, is that in SURREAL the clothing is directly mapped as a texture on top of the naked body, while in 3DPeople the clothing does have its own  geometry.  

As essential as gathering a rich dataset, is the question of what is the most appropriate  geometry representation  for a deep network. In this paper we consider the ``geometry image''   proposed originally in~\cite{Gu_SIGGRAPH2002} and recently used to encode rigid objects in~\cite{SinhaECCV2016,SinhaCVPR2017}. The construction of the geometry image involves two steps, first a mapping of a genus-0 surface onto a spherical domain, and then to a 2D grid resembling an image.  Our contribution here is  on the spherical mapping. We found that existing algorithms~\cite{ChoiSIAM2015,SinhaECCV2016} were not accurate, specially for the elongated parts of the body.  To address this issue we  devise a novel spherical area-preserving parameterization algorithm that combines and extends the FLASH~\cite{ChoiSIAM2015} and the optimal mass transportation methods~\cite{NadeemTVCG2017}. 

Our final contribution consists of designing a  generative network to map input RGB images of a dressed human into his/her corresponding geometry image. Since we  consider $128\times 128\times 3$ geometry images, learning such a mapping is highly complex. We alleviate the learning process through a coarse-to-fine strategy, combined with a series of geometry-aware losses. The full network is trained in an end-to-end manner, and the results are very promising in variety of input data, including both synthetic and real images.  

\section{Related work}
\noindent{\bf 3D Human shape estimation.}  While the problem of localizing the 3D position of the  joints from a single image has been extensively studied  ~\cite{MartinezICCV2017,MehtaSIGGRAPH2017,Moreno_cvpr2017,PavlakosCVPR2017,RogezNIPS2016,SunECCV2018,TomeCVPR2017,VarolCVPR2017,YangCVPR2018,Zhou_2017_ICCV}, the estimation of the 3D body shape has received relatively little attention. This is presumably due to the existence of well-established datasets~\cite{SigalIJCV2010,IonescuPAMI2014}, uniquely annotated with skeleton joints. 

The inherent ambiguity for estimating  human shape from a single view is typically addressed using shape embeddings learned from body scan repositories like SCAPE~\cite{Anguelov_SIGGRAPH2005} and SMPL~\cite{Loper_SIGGRAPH2015}. The body geometry   is   described by a reduced number of pose and shape parameters,  which are  optimized to  match image characteristics~\cite{BalanCVPR2007,BogoECCV2016,LassnerCVPR2017}. 
Dibra \etal~\cite{DibraCVPR2017} are the first in  using a CNN  fed with silhouette images to estimate  shape parameters. In~\cite{TanBMVC2017,TungNIPS2017}  SMPL body parameters are predicted by incorporating  differential renders into the deep network to directly estimate and minimize the error of image features. On top of this, ~\cite{KanazawaCVPR2018} introduces an adversarial loss that penalizes non-realistic body shapes. All these approaches, however, build upon low-dimensional parametric models which are only suitable to  model the geometry of the naked body.

\begin{figure*}[t!]
\centering
\includegraphics[width=1.0\linewidth]{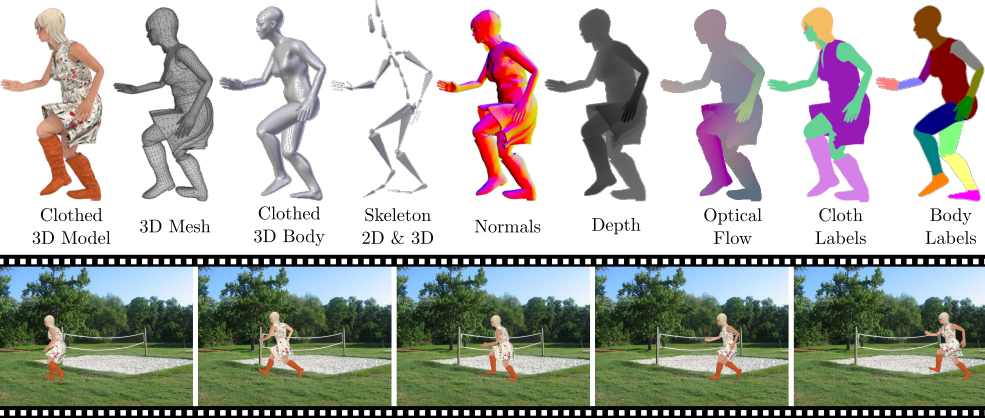}\\
\caption{{\bf Annotations of the 3D People Dataset.} For each of the 80 subjects of the dataset, we generate 280 video sequences (70 actions seen from 4 camera views). The bottom of the figure shows 5 sample frames of the {\em Running} sequence. Every RGB frame is annotated with the information reported in the top of the figure. 3DPeople is the first large-scale dataset with geometric meshes of body and clothes.} 
\label{fig:dataset}
\end{figure*}

\vspace{1mm}
\noindent{\bf Non-parametric representations  for 3D objects.} What is the most appropriate 3D object representation   to train a deep network remains an open question, especially for non-rigid bodies. Standard non-parametric representations for rigid objects  include  voxels~\cite{GirdharECCV2016,YanNIPS2016}, octrees~\cite{TatarchenkoICCV2017,WangSIGGRAPH2017,WangSIGGRAPH2018} and point-clouds~\cite{SuCVPR2017}. ~\cite{SinhaECCV2016,SinhaCVPR2017}   uses  2D embeddings computed with  geometry images~\cite{Gu_SIGGRAPH2002} to represent rigid objects. Interestingly,  very promising results for the reconstruction of non-rigid hands were also reported. DeformNet~\cite{pumarola2018geometry} proposes the first deep model to reconstruct the 3D shape  non-rigid surfaces from a single image. Bodynet~\cite{VarolECCV2018} explores a network that predicts  voxelized human body shape. Very recently,~\cite{NatsumeCVPR2019} introduces a pipeline that given a single  image of a person in frontal position  predicts the body silhouette as seen from different views, and then uses a visual hull algorithm to estimate 3D shape.

\vspace{1mm}
\noindent{\bf Generative Adversarial Networks.} Originally introduced by~\cite{goodfellow2014generative}, GANs have been previously used to model  human body distributions and generate novel images of a person under arbitrary poses~\cite{pumarola2018unsupervised}. Kanazawa~\etal~\cite{KanazawaCVPR2018} explicitly learned the distribution on real parameters of SMPL. DVP~\cite{kim2018deepvideo},  paGAN~\cite{nagano2018pagan} and GANimation~\cite{pumarola2018ganimation} presented models for continuous face animation and manipulation. They have also been applied to edit~\cite{zhou2018talking,song2018talking,vougioukas2018end} and generate~\cite{Duartea2019Wav2Pix} talking faces.

\vspace{1mm}
\noindent{\bf Datasets for body shape analysis.} Datasets are   fundamental in the deep-learning era. While obtaining annotations is quite straightforward for 2D poses~\cite{SappCVPR2013,AndrilukaCVPR2014,JohnsonBMVC2010}, it requires using sophisticated MoCap systems for the 3D case. Additionally, the datasets acquired this way~\cite{SigalIJCV2010,IonescuPAMI2014,IonescuPAMI2014} are mostly indoors. Even more complex is the task of obtaining 3D body shape, which requires expensive setups with muti-cameras  or 3D scanners. To overcome this situation, datasets with synthetically but photo-realistic images have emerged as a  tool to generate massive amounts of training data. SURREAL~\cite{VarolCVPR2017} is the largest and more complete dataset so far, with more than 6 million frames generated by projecting synthetic textures of clothes onto random SMPL body shapes. The dataset is further annotated with body masks, optical flow and depth. However, since clothes are projected onto the naked SMPL shapes just as textures, they can not be explicitly modeled. To fill this gap,  we present the 3DPeople dataset of 3D dressed humans in motion.

\begin{figure*}[t!]
	\centering
	\includegraphics[width=\textwidth]{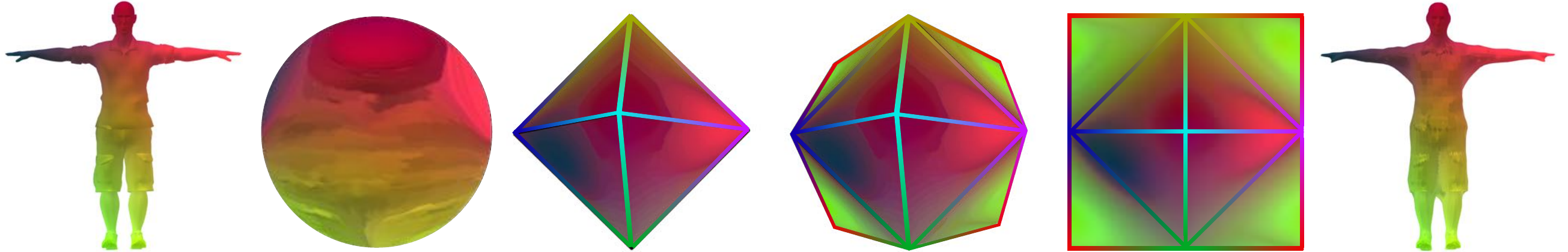}
	\vspace{-11mm}
	\begin{flushleft}
	\small  \hphantom{} \hspace{11mm}
	(a) \hphantom{xxxxxxxaaxxccxx}
	(b) \hphantom{xxxxxxxxcxxxxxx}
	(c) \hphantom{xxxxxxxxxxxxcxxx}
	(d) \hphantom{xxxxxxxxxxxxxxxxx}
	(e) \hphantom{xxxxxxxxxxxxcxx}
	(f) 
	\end{flushleft}
	\vspace{-2mm}
	\caption{{\bf Geometry image representation of  the reference mesh}. (a) Reference mesh in a tpose configuration color coded using the xyz position. (b) Spherical parameterization; (c) Octahedral parameterization; (d) Unwarping the octahedron to a planar configuration; (e) Geometry image, resulting from the projection of the octahedron onto a plane; (f) mesh reconstructed from the geometry image. Colored edges in the ochtahedron and in the geometry image represent the symmetry  that is later exploited by the mesh regressor $\Phi$. }
	\label{fig:gim_reference}
\end{figure*}

\begin{figure}[t!]
\centering
\includegraphics[width=0.8\linewidth]{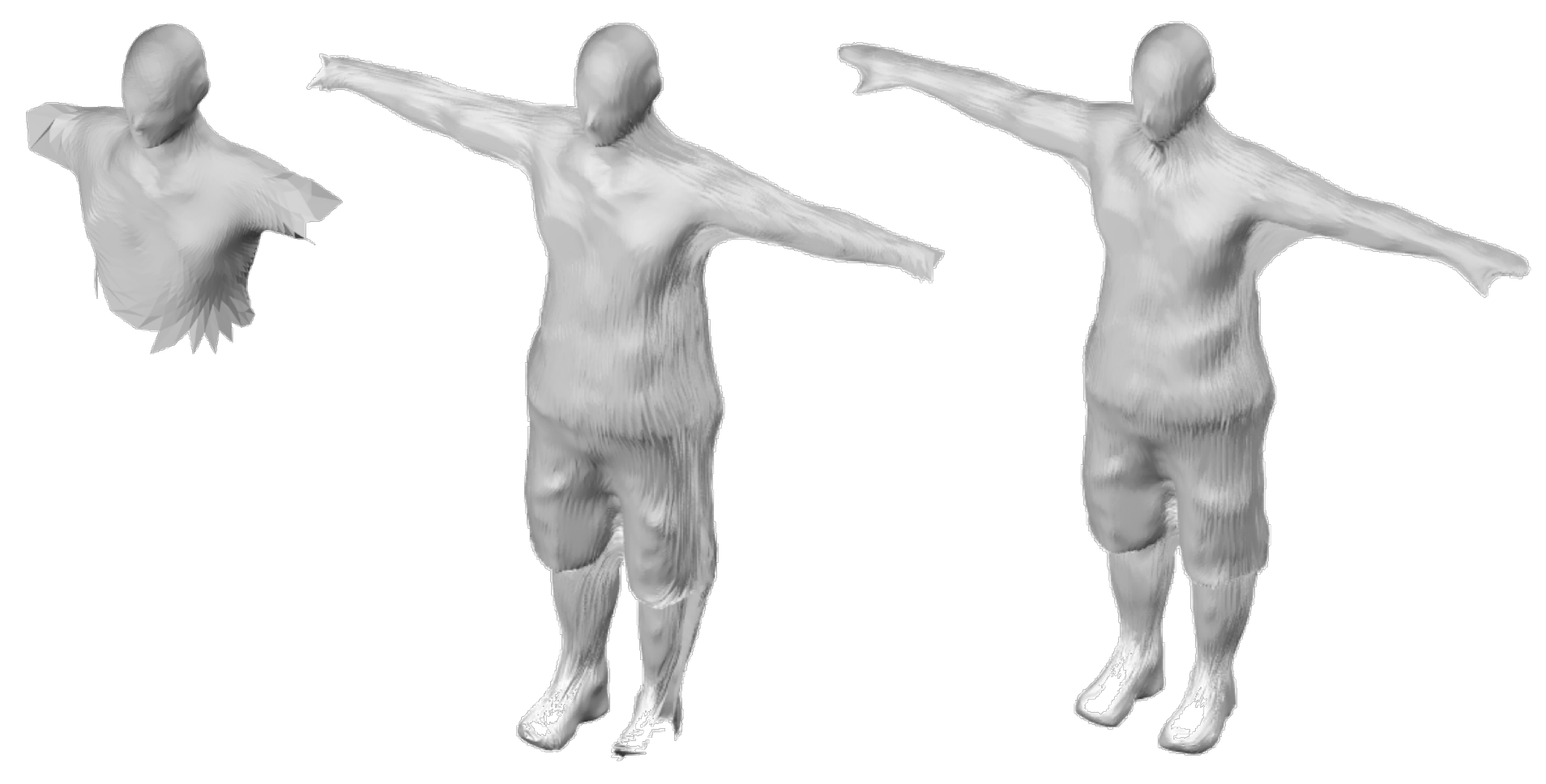}\\
\caption{{\bf Comparison of spherical mapping methods.} Shape reconstructed from a geometry image obtained with three different algorithms. Left: FLASH~\cite{ChoiSIAM2015}; Center:~\cite{SinhaECCV2016}; Right: SAPP algorithm we propose. Note that SAPP is the only method that can effectively recover feet and hands.} 
\label{fig:sph_map_methods}
\end{figure}

\section{3DPeople dataset}
To facilitate future researches, we introduce 3DPeople, the first dataset of dressed humans with specific geometry representation   for the clothes. The dataset in numbers can be summarized as follows: it contains 2.5 Million photorealistic $640\times 480$ images split into 
40 male/40 female performing 70 actions. Every subject-action sequence is captured from 4 camera views, and for each view the texture of the clothes, the lighting direction and the background image are randomly changed. Each frame is annotated with (see Fig.~\ref{fig:dataset}): 3D textured mesh of the naked  and dressed body; 3D skeleton; body part and cloth  segmentation masks; depth map; optical flow; and camera parameters. We next briefly describe the generation process: 

\vspace{1mm}
\noindent{\bf Body  models:} We have generated fully textured triangular meshes    for 80  human characters using Adobe Fuse~\cite{aafuse} and MakeHuman~\cite{aamakehuman}. The distribution of the subjects physical characteristics cover a broad spectrum of body shapes, skin tones and hair geometry (see Fig.~1).  

\vspace{1mm}
\noindent{\bf Clothing models:} Each  subject is dressed with a different outfit including a variety of garments,  combining  tight  and loose clothes. Additional apparel like sunglasses, hats and caps are also included. The final rigged meshes of the body and clothes contain approximately 20K vertices.

\vspace{1mm}
\noindent{\bf Mocap squences:} We gather 70 realistic motion sequences from Mixamo~\cite{aamixamo}. These include human movements with different complexity, from {\em drinking} and {\em typing} actions that produce small body motions to actions like  {\em breakdance} or {\em backflip} that involve very complex  patterns. The mean length of the sequences is of 110 frames. While these are relatively short sequences, they have a large expressivity, which we believe make 3DPeople   also appropriate for exploring action recognition tasks.

\vspace{1mm}
\noindent{\bf Textures, camera, lights and background:} We then use Blender~\cite{aablender} to apply the 70 MoCap animation sequences to each character. Every sequence is rendered from 4 camera views, yielding a total of 22,400 clips. We use a projective camera with a 800 mm focal length and  $640 \times 480$ pixel resolution. The 4 viewpoints correspond approximately to  orthogonal directions  aligned with   the ground. The distance to the subject changes for every sequence to ensure a full view of the body in all frames. The textures of the clothes are randomly changed for every sequence (see again Fig.~1). The illumination is composed of an ambient lighting plus a   light  source at infinite, which direction is   changed per sequence. As in~\cite{VarolCVPR2017} we  render the person on top of a static background image,   randomly taken from the LSUN dataset~\cite{yuARXIV2015}. 

\vspace{1mm}
\noindent{\bf Semantic labels:} For every rendered image, we provide segmentation labels of the clothes (8 classes) and body (14  parts). Observe in Fig.~\ref{fig:dataset}-top-right that the former are aligned with the dressed human, while the body parts are aligned with the naked body. 
 
\begin{figure*}[t!]
	\centering
	\includegraphics[width=\textwidth]{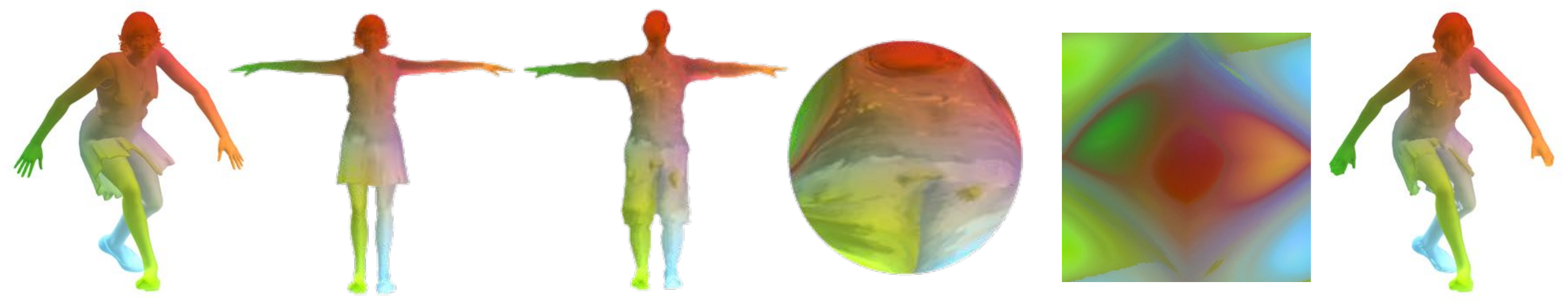}
	\vspace{-12mm}
	\begin{flushleft}
	\small  \hphantom{} \hspace{14mm}
	(a) \hphantom{xxxxxxxxxxccx}
	(b) \hphantom{xxxxxxxxxxxxxxxxx}
	(c) \hphantom{xxxxxxxxxxxxxx}
	(d) \hphantom{xxxxxxxxxxxxxxxx}
	(e) \hphantom{xxxxxxxxxxxxxxx}
	(f) 
	\end{flushleft}
	\vspace{-2mm}
	\caption{{\bf Geometry image estimation for an arbitrary mesh}. (a) Input mesh $\bQ$ in an arbitrary pose color coded using the xyz position of the vertices; (b) Same mesh in a tpose configuration ($\bQ^{\textrm{tpose}}$). The color of the mesh is mapped from $\bQ$; (c) Reference tpose $\bR^{\textrm{tpose}}$. The colors again correspond from those transferred from $\bQ$ through the non-rigid map between $\bQ^{\textrm{tpose}}$ and 
	$\bR^{\textrm{tpose}}$; (d) Spherical mapping of $\bQ$; (e) Geometry image of $\bQ$; (f) Mesh reconstructed from the  geometry image. Note that while being computed through a non-rigid mapping between the two reference poses, the recovered shape is a very good approximation of the input mesh $\bQ$. 
	}
	\label{fig:gim_anymesh}
\end{figure*}

\section{Problem formulation}
Given a single image $\bI \in \mathbb{R}^{H \times W \times 3}$ of a person wearing an arbitrary outfit, we aim at designing a model capable of directly estimating the 3D shape of the clothed body. We represent the body shape through the mesh associated to a  geometry image with $N^2$ vertices $\bX \in \mathbb{R}^{N \times N \times 3}$ where $\bx_i=(x_i,y_i,z_i)$ are the 3D coordinates of the $i$-th vertex, expressed in the camera coordinates system and centered on the root joint $\bx_r$. This representation is a key ingredient of our design, as it maps the 3D mesh to a regular 2D grid structure that preserves the neighborhood relations, fulfilling thus the locality assumption required in CNN architectures. Furthermore, the geometry image representation allows uniformly reducing/increasing the mesh resolution by simply uniformly downsampling/upsampling. This will play an important role in our strategy of designing a coarse-to-fine shape estimation approach. 

We next describe the two main steps of our pipeline: 1) the process of constructing the geometry images, and 2) the deep generative model we propose for predicting 3D shape. 

\section{Geometry image for dressed humans}
The deep network we  describe later will be trained using pairs $\{\bI,\bX\}$ of images and their corresponding geometry image. For creating the geometry images we consider two different cases, one for a reference mesh in a tpose configuration, and another for any other mesh of the dataset. 

\subsection{Geometry image for a reference mesh}\label{sec:gim_ref_mesh}
One of the subjects of our dataset in a tpose configuration is  chosen as a reference mesh. The process for mapping this mesh into a planar regular grid  is illustrated in Fig.~\ref{fig:gim_reference}. It involves the following steps: 

\vspace{1mm}
\noindent{\bf  Repairing the mesh.} Let $\bR^{\textrm{tpose}} \in \mathbb{R}^{N_R \times 3}$ be the reference mesh with $N_R$ vertices in a tpose configuration (Fig.~\ref{fig:gim_reference}-a). We assume this mesh to be a manifold mesh and to be genus-0. Most of the meshes in our dataset, however, do not fulfill these conditions. In order to fix the mesh we follow the heuristic described in~\cite{SinhaECCV2016} which consists of a  voxelization, a selection of the largest connected region of the $\alpha$-shape, and subsequent hole filling using a medial axis approach.  We denote by $\btR^{\textrm{tpose}}$ the repaired mesh.

\vspace{1mm}
\noindent{\bf Spherical parameterization.} Given the repaired genus-0 mesh $\btR^{\textrm{tpose}}$, we next compute the  spherical parameterization $\mS:\btR^{\textrm{tpose}}\rightarrow\bS$ that maps every vertex of $\btR^{\textrm{tpose}}$ onto the unit sphere $\bS$ (Fig.~\ref{fig:gim_reference}-b). Details of the algorithm we use are explained below.
 
 \vspace{1mm}
\noindent{\bf Unfolding the sphere.} The sphere $\bS$  is mapped onto an octahedron   and then cut along edges  to output a flat geometry image $\bX$. Let us formally denote by $\mU:\bS\rightarrow\bX$, and by $\mG^R=\mU \circ \mS:\btR^{\textrm{tpose}}\rightarrow \bX$ the mapping from the reference mesh to the geometry image.
The unfolding process is shown in  Fig.~\ref{fig:gim_reference}-(c,d,e). Color lines in the geometry image  correspond to the same edge in the octahedron, and are split after the unfolding operation. We will later enforce this symmetry constraint when predicting geometry images. 

\subsection{Spherical area-Preserving parameterization}\label{sec:SAPP}
Although there exist several   spherical parameterization schemes (\eg ~\cite{ChoiSIAM2015,SinhaECCV2016}) we found that they tend to shrink the elongated parts of the full body models such as the arms and legs, making the geometry images incomplete (see Fig.~\ref{fig:sph_map_methods}). In this work, we develop a spherical area-preserving parameterization algorithm for genus-0 full body models by combining and extending the FLASH method~\cite{ChoiSIAM2015} and the optimal mass transportation method~\cite{NadeemTVCG2017}. Our  algorithm is particularly advantageous for handling models with elongated parts. The key idea is to begin with an initial parameterization onto a planar triangular domain with a suitable rescaling correcting the size of it. The area distortion of the initial parameterization is then reduced using quasi-conformal composition. Finally, the spherical area-preserving parameterization is produced using optimal mass transportation followed by the inverse stereographic projection.

\begin{figure*}[t!]
\centering
\includegraphics[width=\textwidth]{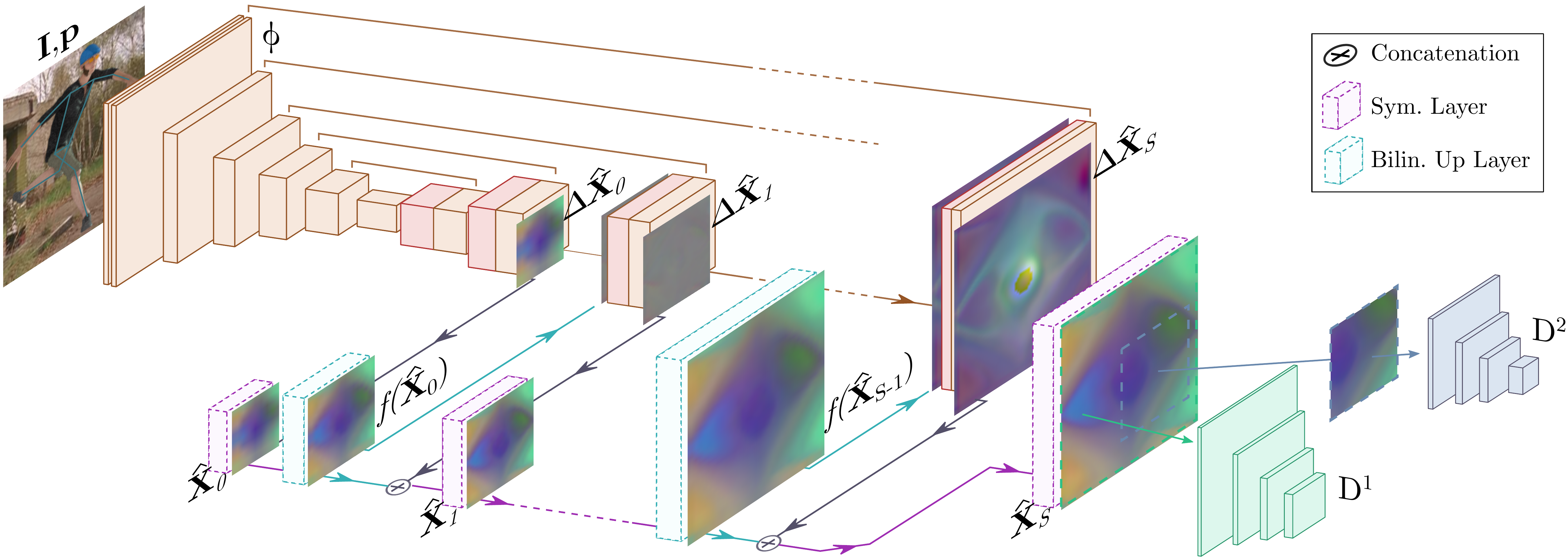}
\caption{{\bf Overview GimNet}. The proposed architecture consists of two main blocks: a multiscale geometry image regressor $\Phi$ and a multiscale discriminator  $D$ to evaluate the local and global consistency of the estimated meshes. 
}
\label{fig:model}
\vspace{-2mm}
\end{figure*}

\subsection{Geometry image for arbitrary meshes}
The approach for creating the geometry image  described in the previous subsection is quite computationally demanding (can last up to 15 minutes for  meshes with complex topology). To compute the geometry image for several thousand  training meshes we have  devised an alternative approach. Let $\bQ  \in \mathbb{R}^{N_Q \times 3}$ be the mesh of any subject of the dataset under an arbitrary pose (Fig.~\ref{fig:gim_anymesh}-a), and let $\bQ^{\textrm{tpose}}  \in \mathbb{R}^{N_Q \times 3}$ be its tpose configuration (Fig.~\ref{fig:gim_anymesh}-b). We assume there is a 1-to-1 vertex correspondence between both meshes, that is,  $\exists\hspace{2mm}\mI:\bQ\rightarrow \bQ^{\textrm{tpose}}$ where $\mI$ is a known bijective function\footnote{This is guaranteed in our dataset, with all meshes of the same subject having the same number of vertices.}.

We then compute  dense correspondences between $\bQ^{\textrm{tpose}}$   and the reference tpose $\btR^{\textrm{tpose}}$, using a nonrigid icp algorithm~\cite{NonrigidICP2019}.  We denote this mapping as $\mN: \bQ^{\textrm{tpose}}\rightarrow\btR^{\textrm{tpose}}$ (see Fig.~\ref{fig:gim_anymesh}-c).  
We can then finally compute the geometry image for the input mesh $\bQ$ by concatenating  mappings:  
\begin{equation}
    \mG^Q=\mG^R \circ \mN \circ \mI: \bQ \rightarrow \bX
\end{equation}
where $\mG^R$ is the mapping from the reference mesh to the geometry image domain estimated in Sec.~\ref{sec:gim_ref_mesh}. It is worth pointing the the nonrigid icp between the pairs of tposes is also highly computationally demanding, but it only needs to be computed once per every subject of the dataset. Once this is done, the geometry image for a new input mesh $\bQ$ can be created in a few seconds.

An important consequence of this procedure is that all geometry images of the dataset will be semantically aligned, that is, every $uv$ entry in $\bX$ will correspond to (approximately) the same semantic part of the model. This will significantly alleviate the learning task of the deep network.

\section{GimNet}
We next introduce {\em GimNet}, our deep generative network to estimate geometry images (and thus 3D shape) of dressed humans from  a single images. An overview of the model is shown in Fig.~\ref{fig:model}. Given the input image, we first extract the 2D joint locations $\bp$ represented as heatmaps~\cite{wei2016cpm,pumarola2018geometry}, which are then fed into a mesh regressor $\Phi(\bI,\bp)$  trained to reconstruct the shape $\bhX$ of the person in $\bI$ employing a geometry image based representation. Due to the high complexity of the mapping (both $\bI$ and $\bhX$ are of size $128 \times 128 \times 3$), the regressor operates in a coarse-to-fine manner, progressively reconstructing meshes at higher resolution. To further enforce the reconstruction to lie on the manifold of anthropomorphic shapes, an adversarial scheme with two discriminators $D$ is applied.

\subsection{Model architecture}
\noindent{\bf Mesh regressor.} Given the input image  $\bI$ and the estimated 2D body joints $\bp$, the mesh regressor $\Phi$ aims to predict the geometry image $\bX$, \ie we seek to estimate the mapping $\mM: \bI, \bp \rightarrow \bX$. Instead of directly learning the  complex mapping $\mM$, we
break the process into a sequence of more manageable steps. $\Phi$ initially estimates a low-resolution mesh, and then progressively increases its resolution (see Fig.~\ref{fig:model}). This coarse-to-fine approach allows the regressor to first focus on the basic shape configuration and then shift attention to  finer  details, while also providing more stability compared to a network that learns the direct mapping.

As shown in Fig.~\ref{fig:gim_reference}-e, the geometry images have symmetry properties derived from unfolding the octahedron into a square, specifically, each side of the geometry image is symmetric with respect to its midpoint. We force this property using a differentiable layer that linearly operates over the edges of the estimated geometry images.

\begin{figure*}[t!]
	\centering
	\includegraphics[width=\textwidth]{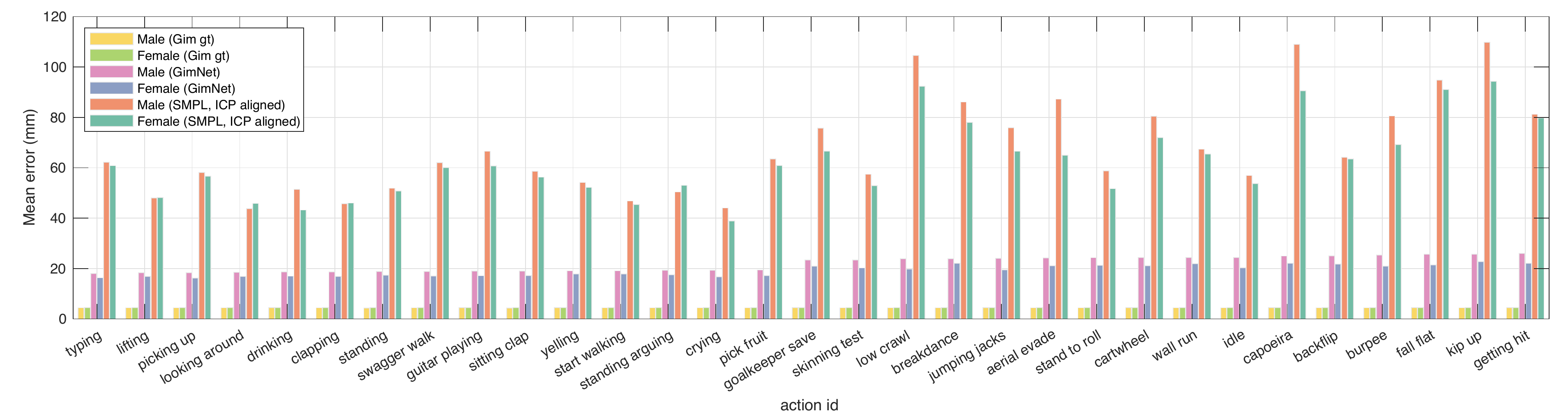}
	\caption{{\bf Mean Error Distance on the test set.} We  plot the results for the 15 worst and 15 best actions. Besides the results of GimNet, we report the results obtained by the ground truth GIM (recall that it is an approximation of the actual ground truth mesh). We also display the results obtained by the recent  parametric approach of~\cite{KanazawaCVPR2018}. The results of this method, however are merely indicative, as we did not retrain the network with our dataset. }
	\label{fig:tableError}
	\vspace{-2mm}
\end{figure*}

\vspace{2mm}
\noindent{\bf Multi-Scale Discriminator.} Evaluating high-resolution meshes poses a significant challenge for a discriminator, as it needs to simultaneously guarantee local and global mesh consistency on very high dimensional data. We therefore use two discriminators with the same architecture, but that operate in different geometry image scales: (i) a discriminator with a large receptive field  that evaluates the shape coherence as a whole; and (ii) a local discriminator that focuses on small patches and enforces the local consistency of the surface triangle faces.

\subsection{Learning the model}

\noindent{\bf 3D reconstruction error.}  We first define a supervised multi-level L1 loss for 3D reconstruction $\mL_{\text{R}}$ as:
\begin{align}
 \mL_{\text{R}} = \mathbb{E}_{\bX \sim \mathbb{P}_r, \bhX \sim \mathbb{P}_g} \frac{1}{S}\sum_{s=1}^{S} \lambda_s \left \| \bX_s - \bhX_s \right \|_{1},
\end{align}
being $\mathbb{P}_r$ and $\mathbb{P}_g$ the real and generated data distribution of clothed human geometry images respectively, $S$ the number of scales, $\bX_s$ the ground-truth reconstruction at scale $s$ and $\bhX_s = \Phi_s(\bI)$ the estimated reconstruction. The error at each scale is weighted by $\lambda_s= \frac{1}{r}$ where $r$ is the ratio between $\bhX_S$ and $\bhX_s$ sizes. During initial experimentation L1 loss reported better reconstructions than mean squared error.

\vspace{2mm}
\noindent{\bf 2D Projection Error.}  To encourage the mesh to   correctly project onto the input image we penalize, at every scale $s$, its projection error $\mL_{\text{P}}$ computed as:
\begin{align*}
&\mL_{\text{P}} = \mathbb{E}_{\bX \sim \mathbb{P}_r, \bhX \sim \mathbb{P}_g} \frac{1}{S}\sum_{s=1}^{S} \lambda_s \left \| \mP(\bX_s) - \mP(\bhX_s) \right \|_{1}
\end{align*}
where $\mP$ is the differentiable projection equation and $\lambda_s$ is calculated as above.

\vspace{2mm}
\noindent{\bf Adversarial loss.} In order to further enforce the mesh regressor $\Phi$ to generate  anthropomorphic shapes we perform a min-max strategy game~\cite{goodfellow2014generative} between the regressor and two discriminators operating at different scales. It is well-known that non-overlapping support between the
true data distribution and model distributions can cause severe training instabilities. As proven by~\cite{roth2017stabilizing, mescheder2018training}, this can be addressed by penalizing the discriminator when deviating from the Nash-equilibrium, ensuring that its  gradients are non-zero orthogonal to the data manifold. Formally, being $ D^{k}$  the $k$\textsuperscript{th} discriminator, the $\mL_{\text{adv}}$ loss is defined as:
\begin{align}
 \sum_{k=1}^{K} \Bigl[ &\mathbb{E}_{\bhX \sim \mathbb{P}_g} [ \log(1-D^{k}(\bhX_S)) ]  \nonumber + \mathbb{E}_{\bX \sim \mathbb{P}_r} \left[ \log(D^{k}(\bX_S)) \right] \\ & + \frac{\lambda_{\text{dgp}}}{2} \mathbb{E}_{\bX \sim \mathbb{P}_r}  ( \| \nabla D^{k}(\bX_S) \|_1^2 \Bigr], 
\end{align}
where $\lambda_{\text{dgp}}$ is a penalty regularization for discriminator gradients, only considered on the true data distribution.

\vspace{2mm}
\noindent{\bf Feature matching loss.}
To improve training stabilization  we penalize higher level features on the discriminators~\cite{wang2018pix2pixHD}. Similar to a perception loss, the estimated geometry image is compared with the ground truth at multiple feature levels of the discriminators. Being $D_{l}^{k}$ the $l$\textsuperscript{th} layer of the $k$\textsuperscript{th} discriminator, $\mL_{\text{F}}$ is defined as:
\begin{align}
\mathbb{E}_{\bX \sim \mathbb{P}_r, \bhX \sim \mathbb{P}_g} \sum_{k=1}^{K}\sum_{l=1}^{L}\frac{1}{N_{l}^{k}}\left \| D_{l}^{k}(\bX_S) - D_{l}^{k}(\bhX_S)\right \|_{1},
\end{align}
where $N_{l}^{k}$ is a weight regularizer denoting the number of elements in the $l$\textsuperscript{th} layer of the $k$\textsuperscript{th} discriminator. 

\vspace{2mm}
\noindent{\bf Total Loss.} Finally, we   to solve the  min-max problem:
\begin{align}
\Phi^\star =\arg \min_{\Phi} \max_{D} \mathcal{L}_{\text{adv}} + \lambda_{\text{R}} \mathcal{L}_{\text{R}} + \lambda_{\text{P}} \mathcal{L}_{\text{P}} + \lambda_{\text{F}} \mathcal{L}_{\text{F}}
\end{align}
where $\lambda_{\text{R}}$, $\lambda_{\text{P}}$ and $\lambda_{\text{F}}$ are the hyper-parameters that control the relative importance of every loss term.  

\begin{figure*}[t!]
	\centering
	\includegraphics[width=\textwidth]{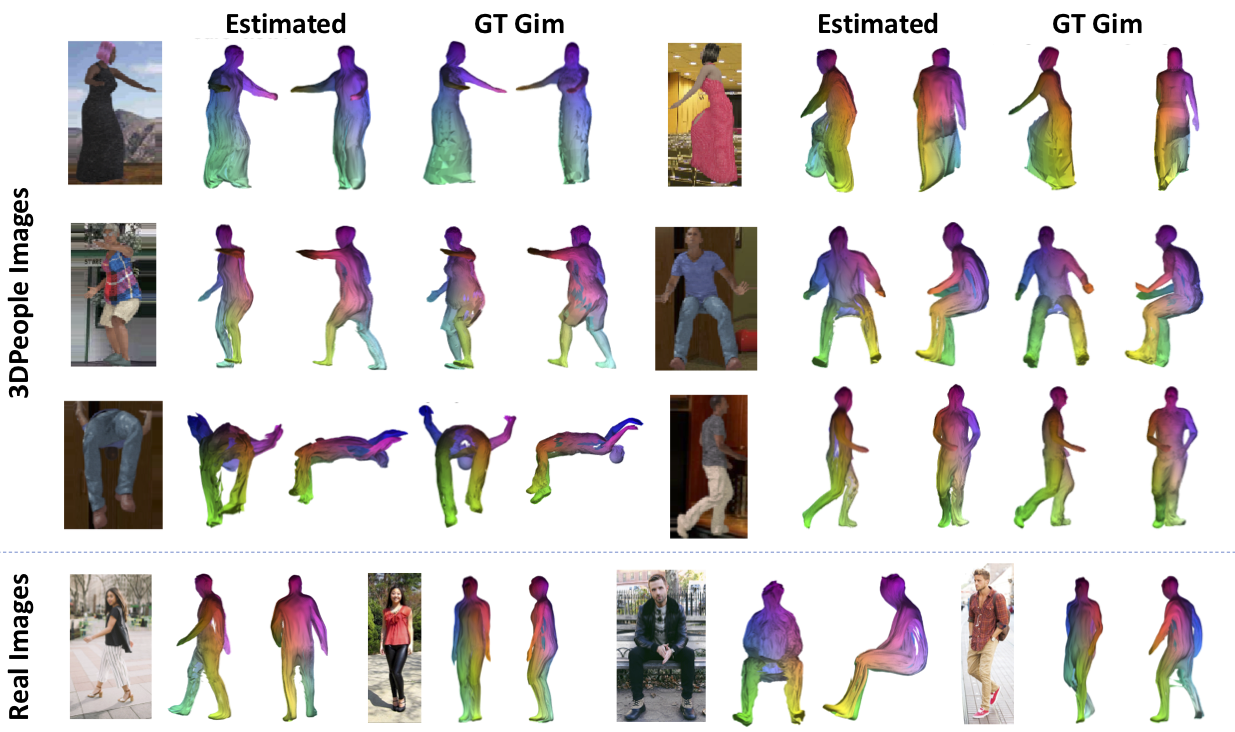}
	\caption{{\bf Qualitative results.} For the synthetic images we plot our estimated results and the shape reconstructed directly from the ground truth geometry image. In all cases we show two different views. The color of the meshes encodes the $xyz$ vertex position.}
	\label{fig:qualitativeResults}
	\vspace{-2mm}
\end{figure*}

\subsection{Implementation details}
For the mesh regressor $\Phi$ we build upon the U-Net architecture~\cite{ronneberger2015u} consisting on an encoder-decoder structure with skip connections between features at the same resolution extended to estimate geometry images at multiple scales. 
Both discriminator networks operate at different mesh resolutions~\cite{wang2018pix2pixHD} but have the same PatchGan~\cite{isola2016image}  architecture mapping from the geometry image $\bX$ to a matrix  $\bY\in\mathbb{R}^{H/8 \times W/8}$, where $\bY[i,j]$  represents the probability of the  patch $ij$ to be close to a real geometry image distribution. The global discriminator evaluates the final mesh resolution at scale $S$ and the local discriminator the down-sampled mesh at scale $S-1$.

The model is trained with 170,000 synthetic images of cropped clothed people resized to $128 \times 128$  pixels and geometry images of $128 \times 128 \times 3$ (meshes with 16,384 vertices) during 60 epochs and $S=4$. As for the optimizer, we use Adam~\cite{kingma2014adam} with learning rate of $2e-4$, beta1 $0.5$, beta2 $0.999$ and batch size $110$. Every $40$ epochs we decay the learning rate by a factor of $0.5$. The weight coefficients for the loss terms  are set to $\lambda_{\text{R}}=20$, $\lambda_{\text{P}}=0.1$, $\lambda_{\text{F}}=10$ and $\lambda_{\text{dgp}}=0.01$.

\section{Experimental evaluation}
We next present   quantitative and qualitative results on synthetic images of our dataset and on images in the wild. 

\vspace{1mm}
\noindent{\bf Synthetic Results}
We   evaluate our approach on 25,000 test images randomly chosen for 8 subjects (4 male/ 4 female) of the test split.  For each test sample we feed   GimNet with the RGB image  and the ground truth 2D pose, corrupted by Gaussian noise with 2 pixel std. For a given test sample, let $\bhY$ be the $N^2\times3$ estimated mesh, resulting from a direct reshaping of its estimated geometry image $\bhX$. Also, let $\bY$ be the ground truth mesh, which does not need to have neither the same number of vertices as $\btY$, nor necessarily the same topology. Since there is no direct 1-to-1 mapping between the vertices of the two meshes we propose using the following metric:  
\begin{equation}
\textrm{dist}(\bhY,\bY)=\frac{1}{2}(\textrm{KNN}(\bhY\rightarrow\bY)+ \textrm{KNN}(\bY\rightarrow\bhY))
\end{equation}
where $\textrm{KNN}(\bhY\rightarrow\bY)$ represents the average Euclidean distance for all vertices of $\bhY$ to their nearest neighbor in $\bY$. Note that $\textrm{KNN}(\cdot ,\cdot)$ is not a true distance measure because it is not symmetric. This is why we compute it bidirectionally. 

The quantitative results are summarized in Fig.~\ref{fig:tableError}. We     report the average error (in mm) of  GimNet for   30  actions  (the 15 with the highest and lowest error). Note that the error of GimNet is bounded between 15 and 35mm. Recall, however, that we do not consider outlier 2D detections in our experiments, but just 2D noise. We also evaluate the error of the ground truth geometry image, as it is an approximation of the actual ground truth mesh. This error is below 5mm, indicating that the geometry image representation does indeed capture very accurately the true shape. Finally, we also provide   the error of the recent parametric approach of~\cite{KanazawaCVPR2018}, that fits    SMPL parameters to the input images. Nevertheless, these results are just indicative, and cannot be directly compared with our approach, as we did not retrain~\cite{KanazawaCVPR2018}. We add them here just to demonstrate the challenge posed by the new 3DPeople dataset. Indeed, the distance error in~\cite{KanazawaCVPR2018} was computed after performing a rigid-icp of the estimated mesh with the ground truth mesh (there was no need of this for GimNet). 

\vspace{1mm}
\noindent{\bf Qualitative Results}
We finally show in Fig.~\ref{fig:qualitativeResults} qualitative results on synthetic images from 3DPeople and real fashion images downloaded from Internet. Remarkably, note how our approach is able to reconstruct long dresses (top row images), known to be a major challenge~\cite{NatsumeCVPR2019}. 
Note also that some of the reconstructed meshes have some spikes. This is one of the limitations of the non-parametric models, that the reconstructions tend to be less smooth than when using  parametric fittings. However, non-parametric models have also the advantage that, if properly trained, can span a much larger configuration space.

\section{Conclusions}
In this paper we have made three contributions  to the problem  of reconstructing the shape of dressed humans: 1) we have presented the first large-scale dataset of 3D humans in action  in which cloth geometry is explicitly modelled; 2) we have proposed a new algorithm to perform spherical parameterizations of elongated body parts, to later model rigged meshes of human bodies as geometry images; and 3) we have introduced an end-to-end network to estimate human body and clothing shape from single images, without relying on parametric models.  While the results we have obtained are very promising, there are still several avenues to explore. For instance, extending the problem to video,  exploring new regularization schemes on the geometry images, or combining segmentation and  3D reconstruction are all  open problems that  can benefit from the  proposed 3DPeople dataset.

\section*{Acknowledgments}
This work is supported in part by an Amazon Research Award, the Spanish Ministry of Science and Innovation under projects HuMoUR TIN2017-90086-R, ColRobTransp DPI2016-78957 and Mar\'ia de Maeztu Seal of Excellence MDM-2016-0656; and by the EU project AEROARMS ICT-2014-1-644271. We also thank NVidia for hardware donation under the GPU Grant Program.

{\small
\bibliographystyle{ieee}
\bibliography{egbib}
}

\end{document}

%% file: newcommands2.tex
\newcommand{\comment}[1]{}

\newcommand{\bp}{\mathbf{p}}

\newcommand{\bx}{\mathbf{x}}

\newcommand{\bI}{\mathbf{I}}

\newcommand{\bQ}{\mathbf{Q}}
\newcommand{\bR}{\mathbf{R}}
\newcommand{\btR}{\tilde{\mathbf{R}}}

\newcommand{\bS}{\mathbf{S}}

\newcommand{\bX}{\mathbf{X}}

\newcommand{\bY}{\mathbf{Y}}
\newcommand{\btY}{\tilde{\mathbf{Y}}}

\newcommand{\bhX}{\hat{\mathbf{X}}}
\newcommand{\bhY}{\hat{\mathbf{Y}}}

\newcommand{\mG}{\mathcal{G}}
\newcommand{\mI}{\mathcal{I}}

\newcommand{\mM}{\mathcal{M}}
\newcommand{\mN}{\mathcal{N}}
\newcommand{\mL}{\mathcal{L}}

\newcommand{\mP}{\mathcal{P}}

\newcommand{\mS}{\mathcal{S}}

\newcommand{\mU}{\mathcal{U}}